\documentclass[sigconf]{acmart}

\usepackage{graphicx}
\usepackage{enumitem}
\usepackage{booktabs}
\usepackage{bbding}
\usepackage{diagbox}
\usepackage{utfsym}
\usepackage{balance}
\usepackage{xcolor}

\AtBeginDocument{%
  }

\copyrightyear{2023} 
\acmYear{2023} 
\setcopyright{acmlicensed}\acmConference[CIKM '23]{Proceedings of the 32nd ACM International Conference on Information and Knowledge Management}{October 21--25, 2023}{Birmingham, United Kingdom}
\acmBooktitle{Proceedings of the 32nd ACM International Conference on Information and Knowledge Management (CIKM '23), October 21--25, 2023, Birmingham, United Kingdom}
\acmPrice{15.00}
\acmDOI{10.1145/3583780.3614782}
\acmISBN{979-8-4007-0124-5/23/10}

\begin{document}

\author{Jingdan Zhang}
\affiliation{%
  \institution{Shanghai Key Laboratory of Data
Science, School of Computer Science,
Fudan University}
  \city{Shanghai}
  \country{China}
}
\email{zhangjd22@m.fudan.edu.cn}

\author{Jiaan Wang}
\affiliation{%
  \institution{School of Computer Science and Technology, \\Soochow University}
  \city{Suzhou}
  \country{China}
}
\email{jawang1@stu.suda.edu.cn}

\author{Xiaodan Wang}
\affiliation{%
  \institution{Shanghai Key Laboratory of Data
Science, School of Computer Science,
Fudan University}
  \city{Shanghai}
  \country{China}
}
\email{xiaodanwang20@fudan.edu.cn}

\author{Zhixu Li}
\affiliation{%
  \institution{Shanghai Key Laboratory of Data
Science, School of Computer Science,
Fudan University}
  \city{Shanghai}
  \country{China}
}
\authornote{Corresponding authors.}
\email{zhixuli@fudan.edu.cn}

\author{Yanghua Xiao}
\affiliation{%
  \institution{Shanghai Key Laboratory of Data
Science, School of Computer Science,
Fudan University}
  \city{Shanghai}
  \country{China}
}
\authornotemark[1]
\email{shawyh@fudan.edu.cn}

\title{AspectMMKG: A Multi-modal Knowledge Graph with Aspect-aware Entities}

\renewcommand{\shortauthors}{Jingdan Zhang, Jiaan Wang, Xiaodan Wang, Zhixu Li, \& Yanghua Xiao}

\begin{abstract}
Multi-modal knowledge graphs (MMKGs) combine different modal data (e.g., text and image) for a comprehensive understanding of entities. Despite the recent progress of large-scale MMKGs, existing MMKGs neglect the multi-aspect nature of entities, limiting the ability to comprehend entities from various perspectives.
In this paper, we construct AspectMMKG, the first MMKG with aspect-related images by matching images to different entity aspects. Specifically, we collect aspect-related images from a knowledge base, and further extract aspect-related sentences from the knowledge base as queries to retrieve a large number of aspect-related images via an online image search engine. 
Finally, AspectMMKG contains 2,380 entities, 18,139 entity aspects, and 645,383 aspect-related images. We demonstrate the usability of AspectMMKG in entity aspect linking (EAL) downstream task and show that previous EAL models achieve a new state-of-the-art performance with the help of AspectMMKG.
To facilitate the research on aspect-related MMKG, we further propose an aspect-related image retrieval (AIR) model, that aims to correct and expand aspect-related images in AspectMMKG.
We train an AIR model to learn the relationship between entity image and entity aspect-related images by incorporating entity image, aspect, and aspect image information. 
Experimental results indicate that the AIR model could retrieve suitable images for a given entity w.r.t different aspects.\footnote{The data is publicly available at \url{https://github.com/theZJD/AspectMMKG}}
\end{abstract}



 % Entity Aspect Linking

\maketitle

\begin{figure*}
  \centering
  \includegraphics[width=0.95\textwidth]{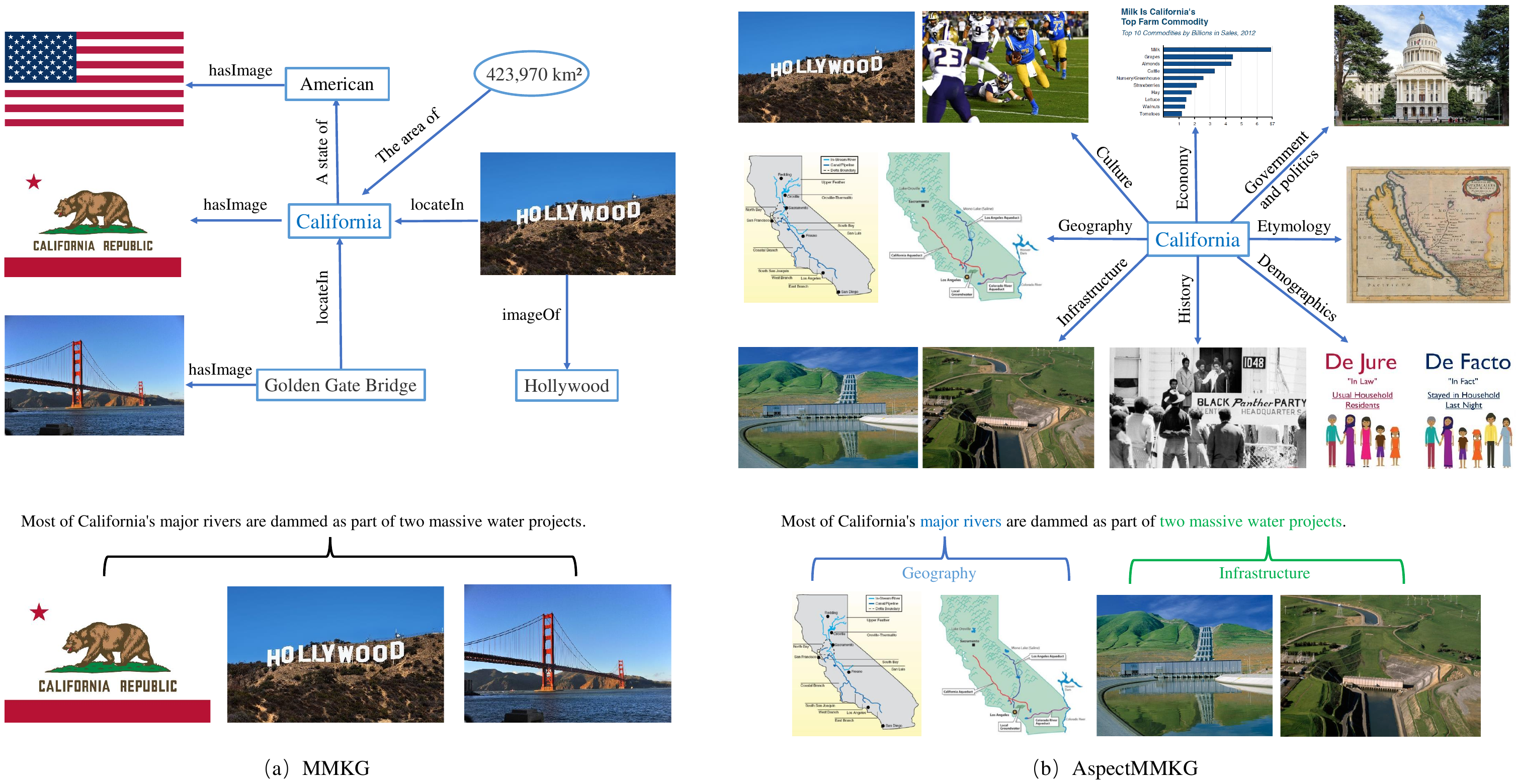}
  \caption{The comparisons between (a) previous MMKG and (b) our AspectMMKG. AsepctMMKG can provide aspect-related images which are highly relevant to the given context.} 
  \label{fig:MMKG}
\end{figure*}

\section{Introduction}
Multi-Modal Knowledge Graphs (MMKGs) use multi-modal data as their source, and describe entities as well as relationships from multiple perspectives~\cite{zhu2022multi}.
Compared with traditional textual knowledge graphs, MMKGs combine multi-modal information of entities organically, thus providing a comprehensive understanding of entities and facilitating the research on multi-modal intelligence.

Though great successes of MMKGs have been achieved in multiple downstream tasks (\emph{e.g.}, image-text matching and visual question answering), existing MMKGs~\cite{zhang2017mmkg,zhao2021boosting,zhang2019vision,imggraph,mmkg,imgpedia} typically attribute a set of indiscriminate images for each entity, which might limit the precise expression of entity images within a specific context. 
For example, the first image that comes into our mind for the entity ``California'' might be a landscape photo. However, when we talk about its location, it could be better to show a map image with its neighboring states.
In addition, when we talk about infrastructure in California, we would like to get some images related to its highways, water conservancy, wind turbines, etc.
Thus, an entity should correspond to different images given different contexts due to its multiple-aspect nature in real applications, where a set of indiscriminate images cannot precisely express its aspect-aware information.

In this paper, we propose a multi-modal knowledge graph with aspect-related entity images, named AspectMMKG, to solve the above issue.
AspectMMKG aims to provide entities with fine-grained aspects and links each entity to different images given different aspects (\emph{i.e.,} aspect-related entity images). In this manner, for a given context, the visual information of an entity can be specified with the help of fine-grained aspects.
Figure~\ref{fig:MMKG}, takes the California entity as an example to show the difference between AspectMMKG and previous MMKG.
In detail, previous MMKG provides images for the entity only with a single relation (\emph{i.e.}, HasImage) while AspectMMKG links different images to the entity with fine-grained aspect labels (\emph{e.g.}, culture, geography, and history). When facing a given context in downstream tasks, previous MMKG cannot effectively help downstream models comprehend the core entity due to the inaccurate or even wrong visual information of matched images. Our AspectMMKG could help downstream models by providing aspect-related images which are highly relevant to the context.

It is non-trivial to construct AspectMMKG due to the difficulties of obtaining aspect-related images for a tremendous number of entities.
Following previous work~\cite{ramsdell,nanni,wang2022knowledge}, we collect entities and their aspect labels from Wikipedia by regarding the section names in each entity's Wikipedia page as its aspect labels. However, the corresponding images for most entities in Wikipedia suffer from sparsity and incompleteness, making it difficult to obtain aspect-related entity images only from Wikipedia. Therefore, we decide to utilize both Wikipedia and an external image search engine as sources to construct AspectMMKG. Specifically, we first pre-define 15 common entity types and extract 2,380 entities with the most views in Wikipedia from these types. Then, we determine the granularity of the entity aspects by using the image-text model (\emph{e.g.}, CLIP\cite{CLIP}) to score the relevance between entity images and different aspects in Wikipedia pages. Next, to enrich the AspectMMKG, for each entity, we extract its aspect-related sentences from Wikipedia as queries to retrieve the aspect-related entity image through an Internet image search engine.
Since the images retrieved from the Internet might involve noise, we further train a discriminative model that can select aspect-related images with high accuracy from the original retrieved results.
Finally, our AspectMMKG contains 2,380 entities, 18,139 entity aspects, as well as 645,383 aspect-related images, where a linkage between an entity and an aspect-related image has an entity aspect label.

We conduct experiments on the entity aspect linking (EAL) downstream task to verify the effectiveness of the constructed AspectMMKG. Experimental results on widely-used eal-dataset-2020~\cite{ramsdell} show that with the help of our AspectMMKG, the original EAL model can achieve a new state-of-the-art performance. Particularly, the fewer text features, the more helpful the images in AspectMMKG are. When there are only two features, the MAP is increased by 34.6\%.

In this paper, our contributions can be summarized as follows:
\begin{itemize}[leftmargin=*,topsep=0pt]
\item To the best of our knowledge, we are the first to construct a multi-modal knowledge graph with aspect-related entity images (\emph{i.e.}, AspectMMKG). Our AspectMMKG contains 2,380 entities, 18,139 entity aspects, and 645,383 aspect-related entity images.
\item To enrich the aspect-related images automatically, we train a discriminative model that can select aspect-related images with high accuracy from the original retrieved results. Compared with the CLIP model, our discriminative model improves recall@10 by 4.4\%.
\item The experimental results on the entity aspect linking (EAL) downstream task show the constructed AspectMMKG can help the EAL baseline achieve a new state-of-the-art performance, indicating its usability and effectiveness.
\end{itemize}

\begin{figure*}
  \centering
  \includegraphics[width=\textwidth]{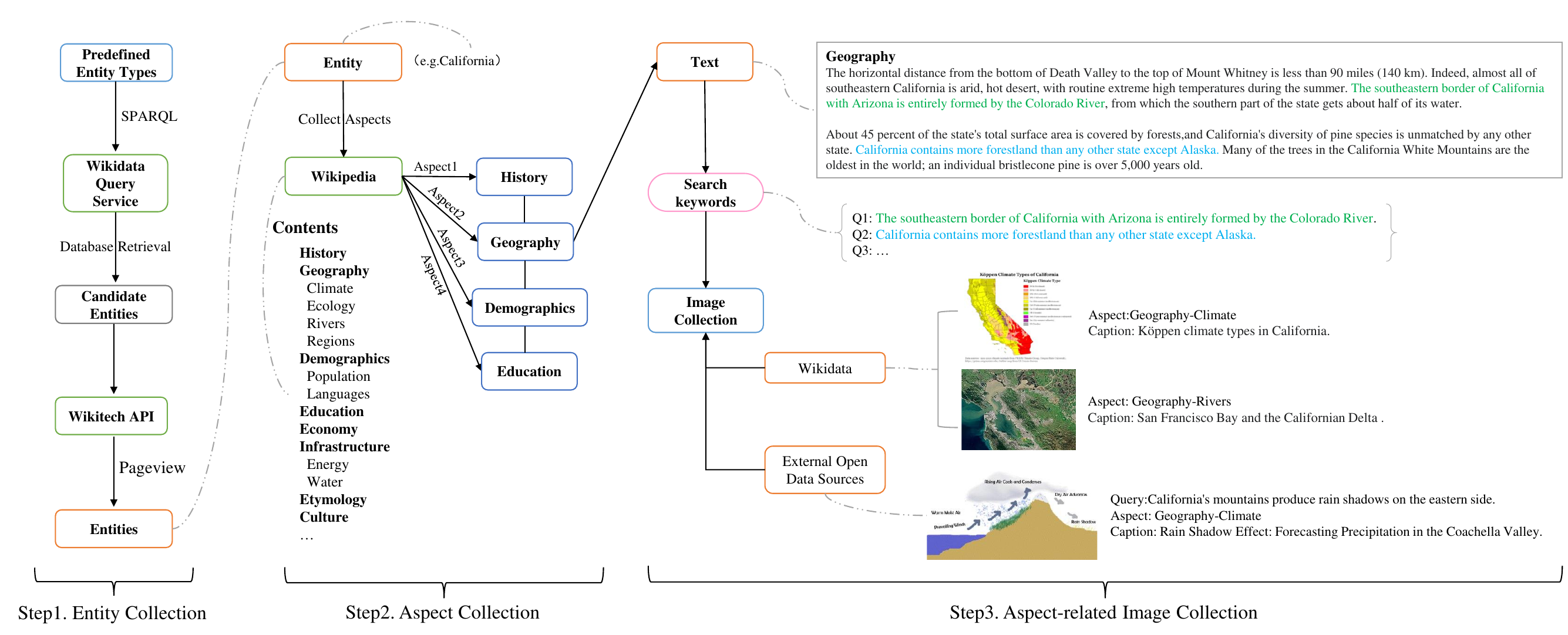}
  \caption{Construction route.}
  \label{fig:route}
\end{figure*}

\section{Data Construction}
We construct AspectMMKG based on Wikipedia and an online image search engine with the three-step process shown in Figure~\ref{fig:route}.
We first obtain 2,380 popular entities from the 15 pre-defined common entity types through Wikidata SPARQL Query (\S~\ref{subsec:2.1}). Then, we extract aspect labels of all these entities from Wikipedia pages (\S~\ref{subsec:2.2}). After that, to enrich the aspect-related images in AspectMMKG, we extract aspect-related sentences from Wikipedia as queries to retrieve aspect-related images via an online image search engine (\S~\ref{subsec:2.3}).
Lastly, we give data statistics (\S~\ref{subsec:2.4}) and quality analysis on AspectMMKG (\S~\ref{subsec:2.5}).

\subsection{Entity Collection}
\label{subsec:2.1}

Following previous work~\cite{ramsdell} and the preliminary observation on Wikipedia, we pre-define 15 entity types which tend to have plenty of aspects.
Table~\ref{table:entity_example} shows all these types, corresponding descriptions, and example entities. Specifically, these types are \textit{Capital and Country}, \textit{Company}, \textit{War}, \textit{Holiday}, \textit{Human (Canada)}, \textit{Human (China)}, \textit{Human (French)}, \textit{Human (UK)}, \textit{Sovereign State}, \textit{State (US)}, \textit{University (Canada)}, \textit{University (UK)}, \textit{University (US)}, \textit{Film (En)}, \textit{Series (En)}.

According to the pre-defined 15 entity types, we next extract popular entities with high pageviews in the following two steps:
(1) We first obtain all entities of these types by leveraging SPARQL statements to retrieve Wikidata Query Service\footnote{\url{https://query.wikidata.org/}}.
(2) We next get the pageview of each entity via Wikitech API\footnote{\url{https://wikitech.wikimedia.org/wiki/Analytics/AQS/Pageviews}} (which can get pageview trends of specific articles or projects) and select popular entities with the top-200 pageviews for each type.
Eventually, we obtain 2,380 entities as the basis of construction AspectMMKG. Note that some entity types, \emph{e.g.}, \textit{State (US)}, might involve less than 200 entities, thus, the total number of collect entities is less than 3000 (15$\times$200). %called head entities

\begin{table*}[]
\resizebox{0.98\textwidth}{!}
{
\begin{tabular}{lll}
\toprule[1pt]
\multicolumn{1}{c}{\textbf{Entity Type}} & \multicolumn{1}{c}{\textbf{Explanation}}          & \multicolumn{1}{c}{\textbf{Example Entities}}                                \\ \midrule[1pt]
Capital and Country              & Country and Country's capital.                  & Tokyo, Paris, France                                    \\ 
Company              & Company or brand                   & Apple Inc, SpaceX, Walmart                                 \\ 
War                  & War                                & Vietnam War, Korean War, Gulf War                          \\ 
Holiday              & Holiday                           & Halloween, Valentine's Day, Thanksgiving                   \\ 
Human (Canada)        & People whose nationality is Canadian.            & Elon Musk, Justin Bieber, James Cameron                    \\
Human (China)         & People whose nationality is Chinese.             & Yao Ming, Jack Ma, Simu Liu                                \\ 
Human (French)        & People whose nationality is French.              & Coco Chanel, Napoleon, Marie Curie                         \\ 
Human (UK)            & People whose nationality is British.             & Elizabeth II, Elton John, Emma Watson                      \\ 
Sovereign State      &  State that has the highest authority over a territory. & Australia, Switzerland, France                             \\ 
State (US)            & US states                          & California, Texas, Florida                                 \\ 
University (Canada) & Canadian universities & McGill University, University of Toronto, Ryerson University        \\ 
University (UK)     & British universities  & University of Oxford, University of Cambridge, University of London \\ 
University (US)     & American universities & Harvard University, Stanford University, Columbia University        \\ 
Film (En)             & English movies                     & Avengers: Endgame, No Time to Die, Spider-Man: No Way Home \\ 
Series (En)           & English TV series                  & WandaVision, Avatar: The Last Airbender, Roman Empire      \\ \bottomrule[1pt]
\end{tabular}
}
\caption{Entity types with the corresponding explanation and example entities.}
\label{table:entity_example}
\end{table*}

\subsection{Aspect Collection}
\label{subsec:2.2}

After collecting entities, we next derive aspect labels of each entity from Wikipedia.
Specifically, given an entity, we extract its aspect labels from the content in its Wikipedia page via a heuristic rule-based method (\emph{i.e.}, parsing HTMLs).
It is worth noting that the content in Wikipedia has a hierarchical structure, leading to the hierarchical aspect labels we collected.
For example, an aspect under the entity ``California'' is ``Geography'', and there are also finer-grained aspects such as ``Regions'' and ``Rivers'' under ``Geography''.

We find that the above-collected entities have some meaningless aspects, such as ``Notes'', ``External links'', ``References'', and ``See also''.
Since these aspects do not contain actual semantic information and cannot reflect information related to entities in specific contexts, we discard these irrelevant aspects.

\subsection{Aspect-related Image Collection}
\label{subsec:2.3}

We decide to collect aspect-related images from both Wikipedia and an online image search engine (\emph{i.e.}, Google) due to the following reasons: (1) Although Wikipedia has high-quality section-related images, the number of images is too limited, according to a comprehensive MMKG survey~\cite{zhu2022multi}, nearly 80\% of English Wikipedia articles have no corresponding images, and only 8.6\% of articles have more than two images; (2) Previous MMKGs typically adopt multi-source data to collect images, for example, the combination of knowledge base and online search engine.

\subsubsection{Collecting Aspect-related Images from Wikipedia}
Given an entity as well as an aspect, we collect the aspect-related images by extracting images belonging to the content in the aspect section. Generally, an image appears in a specific aspect section, the image has a strong relevance to the aspect.

\subsubsection{Collecting Aspect-related Images from Google} 
For an entity, we first extract aspect-related sentences from its Wikipedia page as queries and then use the queries to retrieve images from the Google image search engine.

To extract aspect-related sentences, we locate the aspect with the corresponding section on the entity's Wikipedia page. 
For each aspect of the entity, we extract the sentences from the corresponding section as queries once the sentences contain the entity name.

To collect images from Google, we input each query to the Google image search engine via a python script to retrieve the images and save the first five results (typically, the top result images have a high probability of being images related to the corresponding aspect).
Then, all collected aspect-related images make up our AspectMMKG which contains a total of 2,380 entities, 18,139 entity aspects, and 645,383 aspect-related images.

\subsection{Data Statistics}
\label{subsec:2.4}

As shown in Table~\ref{table:MMKGs}, compared with the existing mainstream MMKGs~\cite{imgpedia,imggraph,mmkg,richpedia,visualsem}, our AspectMMKG is the only MMKG with fine-grained aspect information. Though the number of entities in AspectMMKG is less than others, in this paper, we primarily focus on matching the entities with aspect-related images. Thus, we pay more attention to aspects and the enrichment of aspect-related entity images.
From this perspective, the average number of images per entity in our AspectMMKG is significantly higher than any other MMKGs (271.2 vs. <100), and there are 7.62 aspects per entity on average, facilitating future research with aspect-related MMKGs.

In more detail, the average number of images associated with each entity is 271.2. There is a certain imbalance in the number of each entity. A small number of long-tail entity images are less than other entities. The dataset contains a total of 4,775 aspect labels. These labels can be used To perform knowledge graph reasoning, different aspects of the entity will have a small number of repeated images, which are considered to be error graphs under this aspect, and can be filtered by the graph correction model.

\begin{table*}[]
\resizebox{0.70\textwidth}{!}
{
\begin{tabular}{c|c|c|c|c}
\toprule[1pt]
\textbf{MMKG} & \textbf{Candidate Images} & \textbf{\# Entity} & \textbf{Images / Entity} & \textbf{Aspect} \\ \midrule[1pt]
IMGpedia\cite{imgpedia}   & Wikimedia Commons        & 2.6M   & \textgreater{}5.6 & \usym{2717} \\ 
ImageGraph\cite{imggraph} & Search engine            & 15K    & 55.8              & \usym{2717} \\ 
MMKG\cite{mmkg}       & Search engine            & 15K    & 55.8              & \usym{2717} \\ 
Richpedia\cite{richpedia}  & Search engine, Wikipedia & 29,985 & 99.2              & \usym{2717} \\ 
VisualSem\cite{visualsem}  & Wikipedia, ImageNet      & 89.9K  & 10.4              & \usym{2717} \\ 
AspectMMKG & Search engine, Wikipedia & 2,380  & 271.2             & \usym{2714} \\ \bottomrule[1pt]
\end{tabular}
}
\caption{Data statistics of AspectMMKG and previous MMKGs. ``\# Entity'' denotes the number of entities. ``Images / Entity'' indicates the average number of images per entity. ``Aspect'' represents whether the MMKG provides aspect information.}
\label{table:MMKGs}
\end{table*}

\subsection{Data Quality}
\label{subsec:2.5}
The quality of AspectMMKG is measured by human evaluators. In detail, we randomly select 5 entity types from 15 pre-defined entity types and randomly select an entity from each entity type for evaluation. Then we manually judge whether each aspect-related image is in line with the corresponding aspect, and compute the overall accuracy.
We employ three graduate students as evaluators to make the judgments, and the Fless's Kappa score~\cite{fleiss1971measuring} of the evaluation results is 0.52, indicating a good inter-agreement among our evaluators.
As shown in Table~\ref{table: matching rate}, we find that the aspect-related images of most entities have a decent accuracy (note that there are hundreds of images for an entity), showing the high relevance between aspect-related images and the corresponding aspects in our AspectMMKG.
We also find that the accuracy of ``Dog'' and ``Dragon Boat Festival'' is slightly worse than others. The ``Dog'' entity is a high-level abstract entity and could cover a large number of real-world entities. Thus, the retrieved aspect-related images might be inaccurate.
The image granularity in the Google image search engine of the ``Dragon Boat Festival'' entity is relatively coarse, and there are not enough aspect-related images that reflect the specific semantics of the entity.

\begin{table}[]
\begin{tabular}{c|c}
\toprule[1pt]
\textbf{Entity} & \textbf{Accuracy (\%)} \\ \midrule[1pt]
Dog                  & 72.3                  \\
Apple Inc            & 83.3                  \\ 
Ed Sheeran           & 83.8                  \\ 
Harvard University   & 86.1                  \\ 
Dragon Boat Festival & 64.1                  \\ \bottomrule[1pt]
\end{tabular}
\caption{Human-evaluated accuracy of aspect-related images in AspectMMKG.}
\label{table: matching rate}
\end{table}

\begin{figure}[t]
    \centering
    \includegraphics[width=8cm]{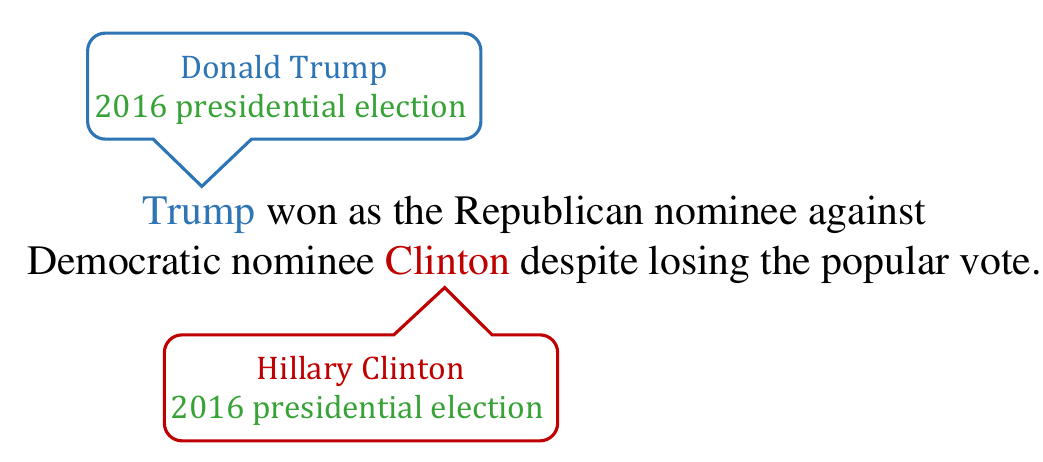}
    \caption{An example of EAL, the green text represents for aspect, and blue and red text represent two entities.}
    \label{fig:EAL}
\end{figure}

\section{Downstream Performance}
\label{sec:3}
We apply AspectMMKG to the entity aspect linking (EAL) downstream task to evaluate the effectiveness of AspectMMKG.

\subsection{Task Definition}
Given a context sentence $S=(s_1,s_2,...s_{|S|})$ ($s_i$ denotes the $i$-th token in $S$) and the involved entity set $\mathcal{E}=\{e_1,e_2,..., e_n\}$, entity aspect linking~\cite{nanni} aims to identify aspect $a_i \in \mathcal{A}$ for each entity $e_i \in \mathcal{E}$ mentioned in the context, where $\mathcal{A}$ denotes pre-defined aspect set.
Figure~\ref{fig:EAL} shows an EAL example, where EAL models should predict the aspect (\emph{i.e.}, ``2016 presidential election'') for the given two entities (\emph{i.e.}, ``Trump'' and ``Clinton'').

\subsection{EAL Framework}
The general framework of EAL models first use a text feature extractor to extract the text features of both the input context $S$ and the aspect sets $\mathcal{E}$:

\begin{equation}
h_s = \operatorname{TextFeature(S=\{s_1,s_2,...s_{|S|}\})}
\end{equation}

\begin{equation}
h_a = \operatorname{TextFeature(\mathcal{E}=\{a_1,a_2,...a_{|a|}\})}
\end{equation}
and then calculate the similarity between the context feature and each aspect feature using different similarity functions: 
\begin{equation}
t_{s_1}, t_{s_2}, ..., t_{s_n} \gets \operatorname{Sim_1(h_s,h_a)}, \operatorname{Sim_2(h_s,h_a)}, ..., \operatorname{Sim_n(h_s,h_a)}
\end{equation}
where $Sim_i$ denotes a similarity function, and there are $n$ functions in total.

After obtaining multiple similarity scores, we input them into a learning-to-rank model to obtain the final EAL prediction.

\begin{equation}
P = \{p_{a_1}, p_{a_2}, ..., p_{a_{|\mathcal{A}|}} \} \gets \operatorname{Learning2Rank}(t_{s_1}, t_{s_2}, ..., t_{s_n})
\end{equation}
where $p_{a_i}$ represents the prediction probability of aspect $a_i \in \mathcal{A}$. The aspect with the highest probability makes the final prediction.

\subsection{Methods}

\begin{table}[t]
\begin{tabular}{c|c|c}
\toprule[1pt]
\textbf{Dataset} & \textbf{entities} & \textbf{intersection} \\ \midrule[1pt]
train-small     & 1000   & 13   \\ 
train-remaining & 106333 & 1046 \\ 
validation      & 1000   & 7    \\ 
test            & 1000   & 9    \\ \bottomrule[1pt]
\end{tabular}
\caption{Our entities and EAL dataset entity analysis.}
\label{table:entity_overlap}
\end{table}

\noindent \textbf{Baseline.} The baseline model is proposed by Ramsdell et al.~\cite{ramsdell}. Specifically, the model computes multiple text similarity scores between context
and different entity aspects, and then fed the similarity scores into a list-wise learning-to-rank model to link the entity mentioned in sentence context to its corresponding aspect.
The following similarity features are adopted in the baseline model: 
\begin{itemize}[leftmargin=*,topsep=0pt]
\item BM25: Using sentence context as a query and each aspect as a document, the model uses BM25 with default parameters as a ranking model.

\item TFIDF: Cosine TF-IDF score between sentence context and aspect part. The baseline uses the TF-IDF variant with TF log normalization and smoothed inverse document frequency.

\item OVERLAP: Number of unique words/entities shared between sentence context and each aspect (no normalization).

\item WORD2VEC: Word embedding similarity between sentence context and each aspect. Word vectors are weighted by their TF-IDF weight. The pre-trained word embeddings are taken from word2vecslim\footnote{\url{https://github.com/eyaler/word2vec-slim}}, a reduced version of the Google News word2vec model.
\end{itemize}

\begin{figure}[t]
    \centering
    \includegraphics[width=8cm]{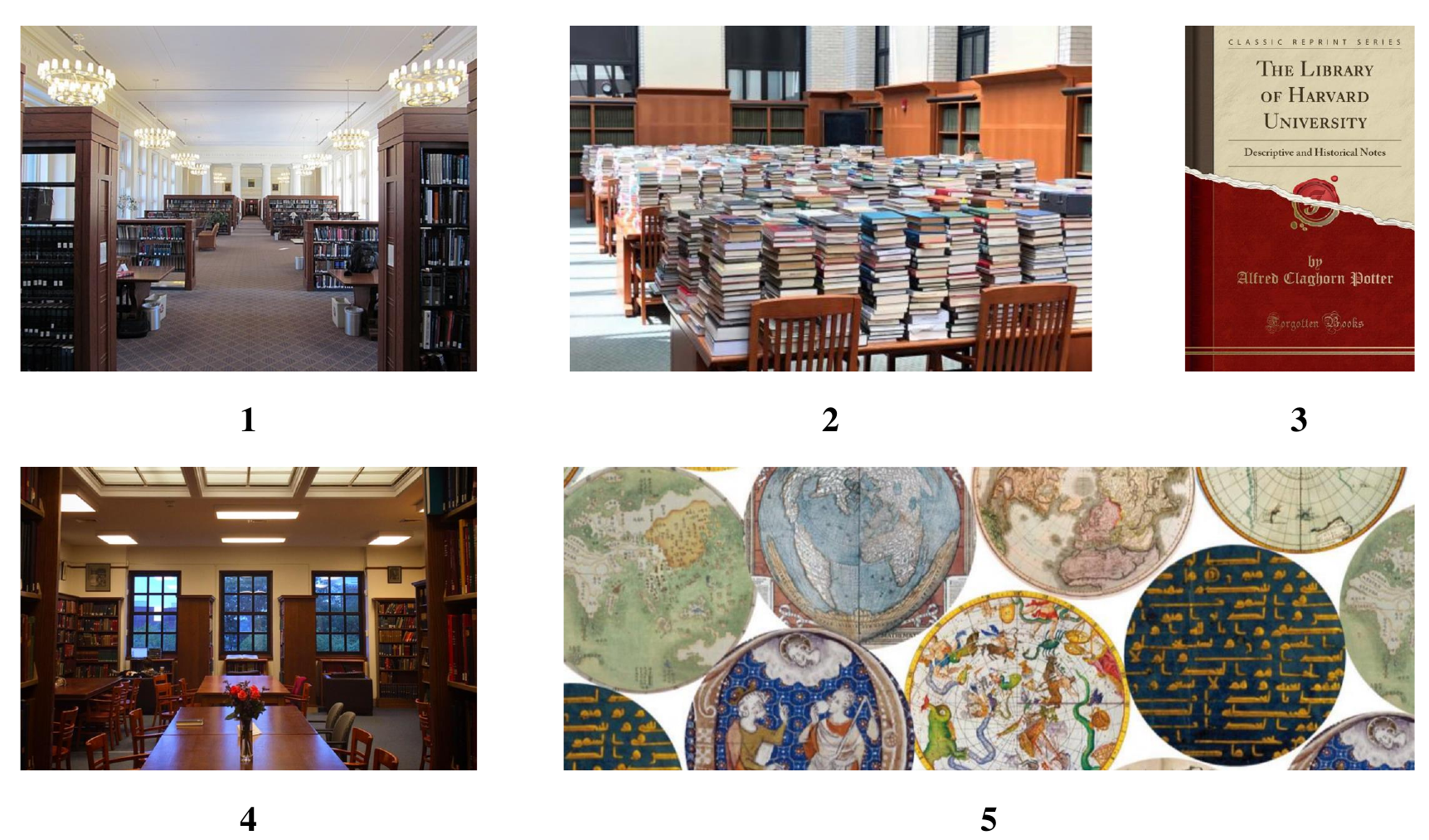}
    \caption{Top five images under the aspect ``Literature'' of entity ``Harvard University'' selected by CLIP.}
    \label{fig:clip-top5}
\end{figure}

\noindent \textbf{Our approach.}
To incorporate the images from AspectMMKG, for an entity and an aspect, considering the balance of aspect image data (there are only about five aspect images with a small number of entities), and the top five images are often more related to aspect tags, we take the images with the top-5 CLIP scores between the images and the aspect to obtain high-relevant images. % that are greater than the average value of the image-text matching score.
A retrieval example is shown in Figure~\ref{fig:clip-top5}. 
Then, for an aspect, we use its high-relevant images to calculate the EAL prediction score. In detail, we use the CLIP text encoder model to encode the sentence context and use the CLIP image encoder to encode the candidate aspect-related images. After that, we calculate the similarity scores between the encoded sentence context and images, and the average similarity score of the five images is used as the EAL score of the given aspect, also named \emph{image feature}.

\subsection{Implementation Details}
We use the machine learning tool Rank-lips\footnote{\url{http://www.cs.unh.edu/~dietz/rank-lips/}} from Ramsdell et al.~\cite{ramsdell} to combine information from multiple features. Rank-lips is a List-wise learning-to-rank toolkit with mini-batched training, using coordinate ascent to optimize for mean-average precision. Mini-batches of 1000 instances are iterated until the training MAP score changes by less than 1\%. To avoid local optima, 20 restarts are used per fold or subset. Z-score normalization is activated.

\subsection{Experimental Setup}

\noindent \textbf{Dataset.} We use the EAL dataset\footnote{\url{http://www.cs.unh.edu/~dietz/eal-dataset-2020/}} provided by Ramsdell et al.~\cite{ramsdell} in the experiment.
Since the EAL dataset only contains the first-level aspect, considering all entities have first-level aspects but not all of them have second-level, or higher-level aspects (\emph{e.g.}, California/Geography and California/Geography/Rivers), we need to unify the granularity of aspects in AspectMMKG. Therefore, we move the images from the higher-level aspect to the first-level aspect.
Besides, the EAL dataset contains a total of 109,392 entities while our AspectMMKG has a total of 2,380 entities, and the overlap between the EAL dataset and AspectMMKG has 1,075 entities, involving 44,574 EAL instances. In our experiment, we use four subsets of the EAL dataset, \emph{i.e.}, the Test dataset, Validation dataset, Train-Small dataset, and Train-remaining dataset.
We do not use the remaining two subsets due to the extremely small number of entities involved.
The number of intersection entities between our AspectMMKG and these datasets is shown in Table~\ref{table:entity_overlap}.

\vspace{0.5ex}
\noindent \textbf{Text Features.} Following Ramsdell et al.~\cite{ramsdell}, there are two aspect features can be used to calculate the similarity between the context and aspect: (a) only the aspect name; (b) the aspect content which is a description text containing aspect name provided by the EAL dataset.
The similarity features combination we use is depicted in Table~\ref{table: similarity}.
Three text similarity features (\emph{i.e.}, BM25, TFIDF and W2Vec.) are used between context and aspect name (since the aspect name might not appear in the context, there is no feature OVERLAP). Four text similarity features (\emph{i.e.}, BM25, TFIDF, Overlap and W2Vec.) are used between context and aspect content. In the end, we use a total of 7 text similarity features. To explore the effect of the image feature conditioned on the different numbers of text features, in addition to selecting all text features, we also randomly select parts of (\emph{i.e.}, 1-6) text features along with the image feature to conduct experiments.

\begin{table}[t]
\resizebox{0.42\textwidth}{!}
{
\begin{tabular}{c|c|c|c|c}
\toprule[1pt]
\textbf{Aspect} & \textbf{BM25} & \textbf{TFIDF} & \textbf{Overlap} & \textbf{W2Vec} \\ \midrule[1pt]
 Aspect Name           & \usym{2714}             & \usym{2714}              &                  & \usym{2714}              \\
Aspect Content        & \usym{2714}             & \usym{2714}              & \usym{2714}                & \usym{2714}              \\ \bottomrule[1pt]
\end{tabular}
}
\caption{The used similarity features in EAL experiments. ``Aspect'' indicates the aspect feature used to calculate similarity is based on aspect name or aspect content.}
\label{table: similarity}
\end{table}

\vspace{0.5ex}
\noindent \textbf{Metric.} We use Mean Average Precision (MAP) as the measurement indicator in our experiments, and the MAP score comprehensively calculates the weighted average accuracy rate of all categories.

\begin{table}[t]
\resizebox{0.48\textwidth}{!}{
\begin{tabular}{c|c|c|c|c|c|c|c}
\toprule[1pt]
\diagbox[dir=NW]{method}{\# Feature} & \textbf{1} & \textbf{2} & \textbf{3} & \textbf{4} & \textbf{5} & \textbf{6} & \textbf{7} \\ \midrule[1pt]
w/o images & 26.4 & 26.7 & 47.8 & 60.6 & 63.5 & 65.4 & 66.8 \\
w/ images   & 41.3 & 61.2 & 64.3 & 65.1 & 65   & 65.5 & 66.9 \\
$\Delta$      & 14.9$^{\ddagger}$ & 34.5$^{\ddagger}$ & 16.5$^{\ddagger}$  & 4.5$^{\ddagger}$  & 1.5$^{\ddagger}$  & 0.1$^{\dagger}$  & 0.1$^{\dagger}$  \\ \bottomrule[1pt]
\end{tabular}
}
\caption{EAL performance in terms of MAP (``\# Feature'' denotes the number of text features inputted to models). $^{\ddagger}$ and $^{\dagger}$ denote that ``w/ images'' method statistically
significant better than the ``w/o images'' method with t-test p < 0.01 and p < 0.05, respectively.}
\label{table: ablation}
\end{table}

\subsection{Results \& Analyses}
The results are shown in Table~\ref{table: ablation}. The horizontal axis represents the number of features input to the ranking model, and the vertical axis represents whether to use AspectMMKG as an auxiliary image feature.
With the help of AspectMMKG, the EAL model achieve new state-of-the-art performance in all scenes, verifying the effectiveness of AspectMMKG.
In addition, it can be seen that the more text features, the less the improvement brought by the images.
Compared with the model involving one text feature, the improvement (\emph{i.e.}, 14.9 MAP score) brought by the image feature is slightly lower than the counterpart (\emph{i.e.}, 34.5 MAP score) of the model involving two text features.
This is because the EAL task is a textual task, images are only used to assist text features to perform EAL, rather than relying solely on the image feature to do this task.
The more features, the higher the accuracy of EAL prediction, which indicates EAL is a complex task and requires multiple features to participate in the learning-to-rank prediction model.

\begin{figure}[t]
    \centering
    \includegraphics[width=8cm]{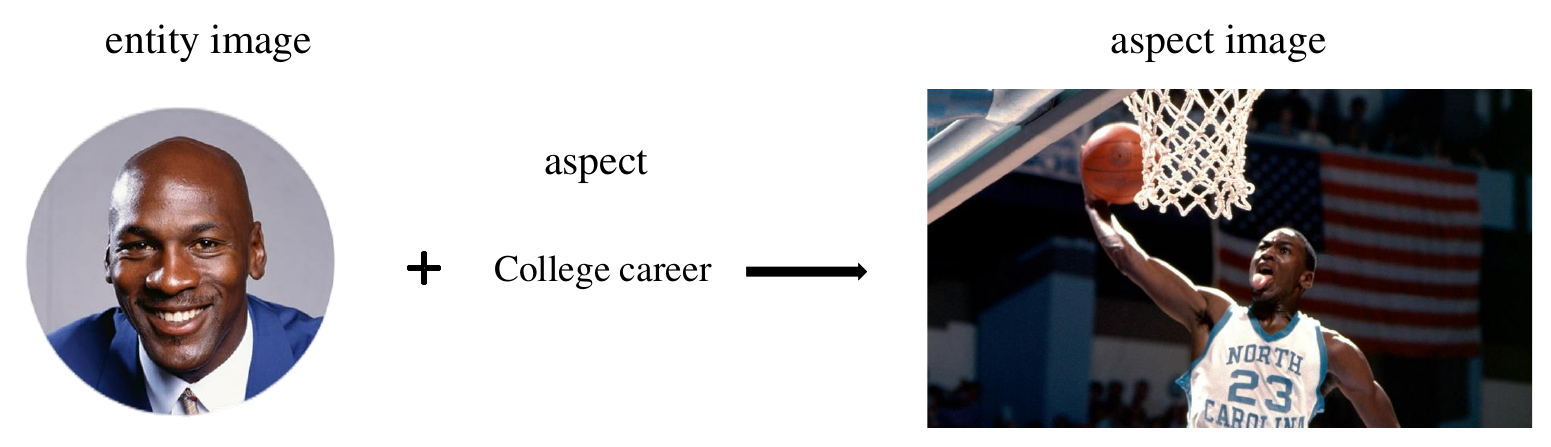}
    \caption{Concept of Aspect-related Image Retrieval}
    \label{fig:subtask-concept}
\end{figure}

\section{Aspect-related Image Retrieval}
To further facilitate the research on aspect-related MMKG, in this section, we propose an aspect-related image retrieval model to retrieve aspect-related images based on the overall entity images.
In detail, for a given entity $e$, the model utilizes its overall image $I_o$ and an aspect label $a \in \mathcal{A}$, to retrieve the aspect-related images from a large number of candidate images $\{I_{c_1}, I_{c_2}, ..., I_{c_k}\}$ (c.f. Figure~\ref{fig:subtask-concept}).

At present, there are many image-text contrast learning models, but there is no image contrast learning model.
We first introduce the application cases of the aspect-related image retrieval model (\S~\ref{subsec:4.1}), and then show the model architecture (\S~\ref{subsec:4.2}).
We next discuss the data used to train the model (\S~\ref{subsec:4.3}) and analyze the model performance (\S~\ref{subsec:4.4}). Finally, we also conduct experiments on the EAL downstream task to show the effectiveness of the retrieval model (\S~\ref{subsec:4.5}).

\begin{figure*}[t]
\centering\includegraphics[height=7cm]
{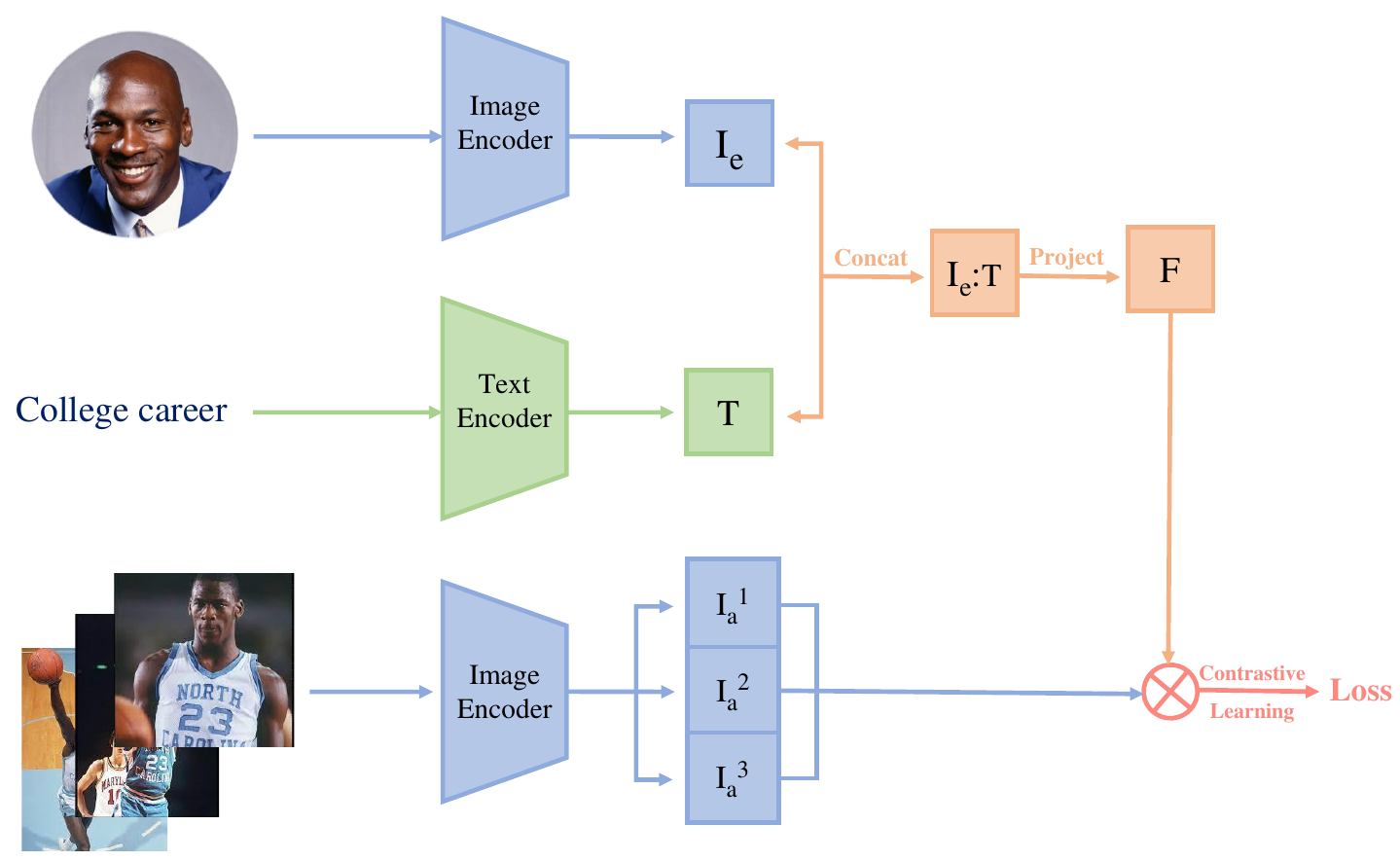}
\caption{The architecture of the aspect-related image retrieval model.}
\label{fig:architecture}
\end{figure*}

\subsection{Application Cases}
\label{subsec:4.1}
The aspect-related image retrieval model can be used to (1) \emph{AspectMMKG correction}: based on existing AspectMMKG, the model can remove incorrect aspect-related images.
(2) \emph{AspectMMKG expansion}: the model can also be used to expand AspectMMKG with more entities as well as aspect-related images in the following situations:
\romannumeral1) For the entities already in AspectMMKG, we can continue to collect some images about the entities via an online image search engine or a knowledge base, and use the model to score the images and divide the collected images into the most suitable aspect nodes;
\romannumeral2) For entities that are not included in AspectMMKG but have aspects in Wikipedia, we still use our method of building AspectMMKG to obtain aspect-related images, and then use the model to filter irrelevant images.
\romannumeral3) For entities that are not included in AspectMMKG and have no aspect in Wikipedia (\emph{e.g.}, long-tail entities), we can search for related images of the entity through an online search engine, but there is no way for these images to know the corresponding aspect labels. We can calculate the similarity between an entity image and different aspects in the AspectMMKG, and use the aspect with the highest similarity as the aspect.

\subsection{Model Architecture}
\label{subsec:4.2}
We use a CLIP-based model for this cross-modal retrieval task.
Figure~\ref{fig:architecture} illustrates the architecture of the model.
The model first leverages the CLIP text encoder to encode the aspect label $a$, and the CLIP image encoder to encode the overall entity image $I_o$ as well as a candidate aspect-related image $I_{c_i}$.

\begin{equation}
\text{H}_{a} \gets \operatorname{CLIPTextEncoder}(a)
\end{equation}
\begin{equation}
\text{H}_{I_o}, \text{H}_{I_{c_i}} \gets \operatorname{CLIPImageEncoder}(I_o), \operatorname{CLIPImageEncoder}(I_{c_i})
\end{equation}

Then we concatenate entity image features ($\text{H}_{I_o}$) and aspect features ($\text{H}_{a})$), and then project to image feature space:
\begin{equation}
\text{H}_{I} = W_I (\text{H}_{I_o} \oplus \text{H}_{a})
\end{equation}
where $\oplus$ means concatenation and $W_I$ denotes trainable parameters.

Next, we use $\text{H}_{I}$ (represents the information of overall image and aspect) and $\text{H}_{I_{c_i}}$ (represents the information of the candidate image) to train the model in a contrastive learning manner:

\begin{equation}
\ell_{i} = -\log \frac{e^{\operatorname{sim}(\mathbf{H}_{I}, \mathbf{H}_{I_{c_i}}^{+}) / \tau}}{\sum_{j=1}^{N} e^{\operatorname{sim}(\mathbf{H}_{I}, \mathbf{H}_{I_{c_j}}^{+}) / \tau}}  
\end{equation}
where $\tau$ is a temperature hyperparameter, and N represents the size of a batch.

\begin{table}[t]
\begin{tabular}{c|c|c|c}
\toprule[1pt]
\textbf{Model} & \textbf{Recall@3} & \textbf{Recall@5} & \textbf{Recall@10} \\ \midrule[1pt]
CLIP           & 16.2             & 26.7             & 52.7              \\ 
AIR             & 19.1             & 30.5             & 57.1              \\ 
$\Delta$          & 2.9              & 3.8              & 4.4              \\ \bottomrule[1pt]
\end{tabular}
\caption{Aspect-related image retrieval results in terms of Recall@$k$ ($k\in\{3,5,10\}$)}
\label{table: recall}
\end{table}

\subsection{Retrieval Data}
\label{subsec:4.3}
The structure of the training sample is a triplet of $\langle$overall entity image, aspect, aspect-related image$\rangle$. We collect such training samples from AspectMMKG.
Given an entity, to collect the overall image, we use the CLIP model to score relevance between each image w.r.t the entity from AspectMMKG and the entity name, and take the image with the highest relevance as its overall image.
To further collect the aspect-related images, we score the relevance between each aspect-related image from AspectMMKG and the aspect label, and then use the images with the top-3 relevances as the ground truth of the aspect-related images. 

Finally, we collect 46,779 aspect-related image retrieval samples, and divide them into the training set, the validation set and the test set with the proportion of 8 (37453) : 1 (4663) : 1 (4663).

\begin{table}[t]
\begin{tabular}{c|c|c}
\toprule[1pt]
\textbf{Rank} & \textbf{CLIP}           & \textbf{AIR}             \\ \midrule[1pt]
1             & \textbf{Notable people} & \textbf{Notable people} \\ 
2             & Geography               & Sports                  \\ 
3             & Sports                  & Education               \\ 
4             & History                 & \textbf{Notable people} \\ 
5             & Government              & Media                   \\ 
6             & Demographics            & Government              \\ 
7             & Economy                 & History                 \\ 
8             & History                 & Sports                  \\ 
9             & Government              & Media                   \\ 
10            & \textbf{Notable people} & Parks and recreation    \\ \bottomrule[1pt]
\end{tabular}
\caption{A retrieval instance on the test set, the bold text is the distribution of ground truth.}
\label{tabel: case}
\end{table}

\subsection{Results \& Analyses}
\label{subsec:4.4}
We evaluate the aspect-related image retrieval model via (1) automatic evaluation and (2) human evaluation.

\vspace{0.5ex}
\noindent \textbf{Automaic Evaluation.} For an aspect (\emph{e.g.}, urban structure) of an entity (\emph{e.g.}, Sydney), we use all the images under the entity from the test set as candidate aspect-related images, and sort them via the aspect-related image retrieval model, and evaluate the results in terms of recall@3, recall@5 and recall@10.\footnote{Note that the value of Hits@k is equal to Recall@k under our setting.} 
We compare our model with vanilla CLIP, the experimental results are shown in Table~\ref{table: recall}, where we can find that our model can better rank the correct aspect-related images among the whole entity images. Besides, the model has the ability to recognize multi-modal information with fine-grained semantics ``aspect''. 
Additionally, we show a case from the test set in Table~\ref{tabel: case}, with the aspect labels of retrieval images of our model as well as vanilla CLIP on the aspect ``Notable people'' of the entity ``Las Vegas''.

\vspace{0.5ex}
\noindent \textbf{Human Evaluation.}
Based on the AspectMMKG we built, given a certain entity and an aspect, we use our model to calculate the probability of each aspect-related image from AspectMMKG is in line with the aspect label to see if the model can predict the correct image with a high probability, and wrong images with a low probability.
We randomly select 20 entities and evaluate the model by employing three graduate students as evaluators to score the retrieved top-$k$ images for each entity, and calculate the proportion (denotes as P@$k$) of correct images in the top-$k$ retrieval results.

The human evaluation results are shown in Table~\ref{tabel: eval result}.
We can see that because the overall quality of AspectMMKG is relatively high when doing retrieval tasks on it, there are more correct images in the top-K images.
Compared with the CLIP model, our retrieval model achieves better performance in terms of all metrics.

\begin{table}[]
\begin{tabular}{c|c|c|c|c}
\toprule[1pt]
\textbf{Model} & \textbf{P@5} & \textbf{P@10} & \textbf{P@15} & \textbf{P@20} \\ \midrule[1pt]
CLIP           & 76.4        & 80.8         & 83.3         & 84.3         \\ 
AIR            & 84.2        & 86.9         & 88.9         & 88.6         \\ \bottomrule[1pt]
\end{tabular}
\caption{The proportion of correct images among the top-K retrieved images.}
\label{tabel: eval result}
\end{table}

\subsection{EAL Performance}
\label{subsec:4.5}
The effectiveness of the EAL downstream task could also verify the effectiveness of our model, and we leverage the aspect-related image retrieval model to conduct experiments on EAL downstream tasks as described in Section~\ref{sec:3}.
To adapt our model to EAL, we first use our text encoder to encode the aspect, and use our image encoder to encode the candidate aspect-related images. Then, we calculate the similarity scores between aspect labels and candidate aspect-related images, and take the first five images as the EAL aspect-related images.
Next, we use the CLIP text encoder and image encoder to encode the sentence context and the EAL aspect-related images respectively.
We calculate the similarity score between the context and each EAL aspect-related image.
Finally, we regard the average of the five similarity scores as the prediction probability of the entity aspect described in the context.
Table~\ref{table: update ablation} shows the results, our model is better at capturing aspect images than the baseline model (\emph{i.e.}, vanilla CLIP), indicating its superiority.
\begin{table}[t]
\begin{tabular}{c|c|c|c|c|c|c|c}
\toprule[1pt]
\textbf{Type} & \textbf{1} & \textbf{2} & \textbf{3} & \textbf{4} & \textbf{5} & \textbf{6} & \textbf{7} \\ \midrule[1pt]
w/o images & 26.4 & 26.7 & 47.8 & 60.6 & 63.5 & 65.4 & 66.8 \\ 
w images   & 44.4 & 61.3 & 64.3 & 65.2 & 65.4 & 66.7 & 67.1 \\ 
$\Delta$      & 18.0   & 34.6 & 16.5 & 4.6  & 1.9  & 1.3  & 0.3 \\ \bottomrule[1pt]
\end{tabular}
\caption{EAL performance in terms of MAP by using our aspect-related image retrieval model.}
\label{table: update ablation}
\end{table}

\section{Related Work}
\subsection{Multi-modal Knowledge Graph}
The existing multi-modal knowledge graphs are listed as follows:
(1) IMGPEDIA~\cite{imgpedia} is a vast linked dataset that combines visual information from the WIKIMEDIA COMMONS dataset. It includes descriptors for the visual content of 15 million images, 450 million visual-similarity relations between these images, links to image metadata from DBPEDIA COMMONS, and connections to DBPEDIA resources associated with each image.
(2) ImageGraph~\cite{imggraph} is a visual-relational knowledge graph (KG) that features a multi-relational graph where entities are linked to images. It contains 1,330 relation types, 14,870 entities, and 829,931 images collected from the web.
(3) MMKG~\cite{mmkg} is a collection of three knowledge graphs that incorporate both numerical features and image links for all entities. It also includes alignments between pairs of KGs, providing a comprehensive resource for various tasks.
(4) Richpedia~\cite{richpedia} aims to create a comprehensive multi-modal knowledge graph by associating diverse images with textual entities in Wikidata. It establishes RDF links (visual semantic relations) between image entities based on hyperlinks and descriptions in Wikipedia. Richpedia offers a web-accessible faceted query endpoint and serves as a valuable resource for knowledge graph and computer vision tasks, such as link prediction and visual relation detection.
(5) VisualSem~\cite{visualsem} is a high-quality knowledge graph (KG) that includes nodes with multilingual glosses, multiple illustrative images, and visually relevant relations. It also provides a neural multi-modal retrieval model capable of using images or sentences as inputs to retrieve entities in the KG. This retrieval model can be seamlessly integrated into any neural network-based model pipeline.
(6) PKG~\cite{li-etal-2022-multi-modal} is an MMKG for classical Chinese poetry. This MMKG first collects poetry data from Online resources. Then, inspired by the promising performances of large language models~\cite{wang-etal-2023-towards-unifying,wang2023chatgpt,wang2022survey,wang2022clidsum,wang2023zero,shi2022layerconnect}, PKG uses a text-to-image generative model to collect images.

Different from previous MMKGs, we propose AspectMMKG with aspect-related entity images, and an entity in AspectMMKG is linked to different images given different aspects.

\subsection{Aspect Information in KG}

In recent years, researchers propose the definition of entity aspects and propose some methods to identify, extract and utilize the entity aspects to support tasks such as search, classification, and recommendation of more accurate and fine-grained entities. Fetahu et al.~\cite{fetahu} is the first to formalize the research on entities. Their work uses news articles to enrich Wikipedia chapters. Although the word "aspect" is not explicitly used in their paper, they believe that each part represents a different topic. But it only focuses on the most salient entities in the text. On the contrary, Nanni et al.~\cite{nanni} build a system that works for any type of entity mentioned in context. And under the condition that entities have different aspects, the task of entity aspect linking is proposed and solved, i.e., given a context paragraph and a mentioned entity, the task of entity linking is refined according to the aspects of the entity it refers to. Banerjee et al.~\cite{banerjee} and Sauper and Barzilay\cite{sauper} use a combination of retrieval and abstract summarization to predict content to populate new Wikipedia pages. Their focus is to generate content suitable for new pages. Reinanda et al.~\cite{reinanda} introduce the definition of entity "aspect", using the aspect catalog of entity types, and using aspect features to identify related documents of long-tail entities. Arnold et al.~\cite{arnold2019,arnold2020} use chapters of Wikipedia articles to learn how to segment articles into different topics and identify answer paragraphs to biomedical questions. Balasubramanian and Cucerzan~\cite{balasubramanian} build topic pages for popular person entities and use query logs to derive aspects of the entities.
In addition, some researchers~\cite{pantelmining,ligenerating,yinbuilding,reinanda,taneva} extract information from large-scale query logs or perform entity analysis to generate salient content about entities. However, these methods are often dependent on the corpus involved in the study and cannot be effectively extended to rare long-tail entities. Ramsdell et al~\cite{ramsdell}. use the aspect catalog to define aspects for each entity instance. Unlike previous work, we use sections from Wikipedia as sources for aspects instead of query logs. The aspect table of contents is defined using Wikipedia's chapter hierarchy, where the names of the aspects are derived from the chapter titles.
Complementing the work of Nanni et al~\cite{nanni}., Ramsdell et al~\cite{ramsdell}. provide a large-scale test set and provide strong baseline results, and a fine-grained feature set to stimulate more research in this area.

We follow the approach of Ramsdell et al~\cite{ramsdell}., using Wikipedia's chapter hierarchy to define the aspect table of contents, where the names of the aspects are derived from the chapter titles. But the difference is that we have derived multiple levels of aspects. For aspects of different granularities, we use the file system hierarchy to save aspects. Different from previous work on entity aspect linking of plain text features, we employ features that incorporate multi-modal information. It is used to assist text features to complete the task of linking entities.

\section{Conclusion}
In this paper, we construct an MMKG, named AspectMMKG, with a fine-grained semantic ``aspect'' of Entities. We verify the effectiveness of AspectMMKG on the entity aspect linking downstream task. To facilitate the research on aspect-related MMKG, we further train an aspect-related image retrieval model that can be used to correct and expand AspectMMKG.

\section{Acknowledgements} 
This work is supported by the National Natural Science Foundation of China (No.62072323), Shanghai Science and Technology Innovation Action Plan (No. 22511105902, 22511104700), Shanghai Municipal Science and Technology Major Project (No.2021 SHZDZX0103), and Science and Technology Commission of Shanghai Municipality Grant (No. 22511105902).

\bibliographystyle{ACM-Reference-Format}
\balance
\bibliography{sample-base}

\end{document}